\documentclass[runningheads]{llncs}
\usepackage{graphicx}
\usepackage{tikz}
\usepackage{comment}
\usepackage{amsmath,amssymb} % define this before the line numbering.
\usepackage{color}
\usepackage{xcolor}

\usepackage[accsupp]{axessibility}  % Improves PDF readability for those with disabilities.

\usepackage{subcaption}
\usepackage{multirow, multicol}
\usepackage{nicefrac}
\usepackage{microtype}
\usepackage{bm}
\usepackage{booktabs}
\usepackage[symbol]{footmisc}
\usepackage[colorlinks,linkcolor=red]{hyperref}

\usepackage[capitalize]{cleveref}
\crefname{section}{Sec.}{Secs.}
\Crefname{section}{Section}{Sections}
\Crefname{table}{Table}{Tables}
\crefname{table}{Tab.}{Tabs.}

\begin{document}
% \linenumbers
\pagestyle{headings}
\mainmatter
\def\ECCVSubNumber{4349}  % Insert your submission number here

\title{Revisiting the Critical Factors of Augmentation-Invariant Representation Learning} % Replace with your title

\begin{comment}
\titlerunning{ECCV-22 submission ID \ECCVSubNumber} 
\authorrunning{ECCV-22 submission ID \ECCVSubNumber} 
\author{Anonymous ECCV submission}
\institute{Paper ID \ECCVSubNumber}
\end{comment}
%******************

% \begin{comment}
\titlerunning{Revisiting the Critical Factors of AIRL}
\author{Junqiang Huang\textsuperscript{$\dagger$} \and Xiangwen Kong\textsuperscript{$\dagger$} \and Xiangyu Zhang} 
%\footnotemark[1]{equal contribution.}

 \authorrunning{J. Huang et al.}
\institute{MEGVII Technology, Beijing, China \email{\{huangjunqiang,kongxiangwen,zhangxiangyu\}@megvii.com}\\
}


% \end{comment}
%******************
\maketitle

\renewcommand{\thefootnote}{\fnsymbol{footnote}}
\footnotetext[2]{Equal Contribution}
\renewcommand{\thefootnote}{}
\footnotetext{Code: \url{https://github.com/megvii-research/revisitAIRL}}

\renewcommand*{\thefootnote}{\arabic{footnote}}

\begin{abstract}

We focus on better understanding the critical factors of augmentation-invariant representation learning. We revisit \mbox{MoCo v2} and BYOL and try to prove the authenticity of the following assumption: different frameworks bring about representations of different characteristics even with the same pretext task. We establish the first benchmark for fair comparisons between \mbox{MoCo v2} and BYOL, and observe: (i) sophisticated model configurations enable better adaptation to pre-training dataset; (ii) mismatched optimization strategies of pre-training and fine-tuning hinder model from achieving competitive transfer performances. Given the fair benchmark, we make further investigation and find asymmetry of network structure endows contrastive frameworks to work well under the linear evaluation protocol, while may hurt the transfer performances on long-tailed classification tasks. Moreover, negative samples do not make models more sensible to the choice of data augmentations, nor does the asymmetric network structure. We believe our findings provide useful information for future work.

% \keywords{Augmentation-Invariant Representation Learning}
\end{abstract}

\section{Introduction}

Recently, with the advancement of research on pretext tasks \cite{exemplar_cnn,context_prediction,rotation,jigsaw,cpc,inst_disc}, self-supervised learning (SSL) presents extraordinary potential in computer vision, pushing the frontier of transfer learning. The effectiveness of self-supervised learned representations has been empirically verified. Compared to supervised pre-training counterparts, MoCo series \cite{moco,mocov2,mocov3} achieves comparable or even better performances on object detection, semantic segmentation, etc. Moreover, under the linear evaluation protocol on ImageNet \cite{imagenet} (an often used evaluation metric for SSL), BYOL \cite{byol} and SwAV \cite{swav} have largely shrunk the gap with supervised learning.

Among various pretext tasks, one of the most promising ways is to pull together the positive sample pairs (different augmented views of the same image), which enables the model to learn augmentation-invariant representations. The simplicity of this pretext task also brings about a notorious problem: without careful design, the model will collapse to a trivial solution that all images are mapped to a constant vector, resulting in useless representations. To avoid this collapse, contrastive methods like \mbox{MoCo} impose regularization by pushing away the negative sample pairs (different images), while BYOL develops the asymmetric siamese network with a stop-gradient operation. Though sharing the same pretext task, \mbox{MoCo v2} and BYOL show different results of linear classification and transfer learning. As reported in \cite{byol,simsiam}, BYOL has higher linear accuracy, while \mbox{MoCo v2} presents better transferability. Given this observation, it is natural to assume different frameworks bring about representations of different characteristics. 

To prove or disprove the above assumption, it is essential to build the benchmark for fair comparison between contrastive frameworks and BYOL. Since \mbox{MoCo v2} shares many similarities with BYOL, which is convenient to perform controlled experiments, we choose it as the representative of contrastive methods. We aim to study the experimental impact of the following variables on augmentation-invariant representation learning: model configurations (i.e., network architecture, symmetry of training loss, etc.), combination of data augmentations, and optimization strategies. The evaluation criteria consist of linear classification accuracy and transfer performances of typical downstream tasks. Our efforts and contributions will be described next.

\begin{figure*}
  \centering
  \begin{subfigure}{0.24\linewidth}
    \includegraphics[width=1.0\textwidth]{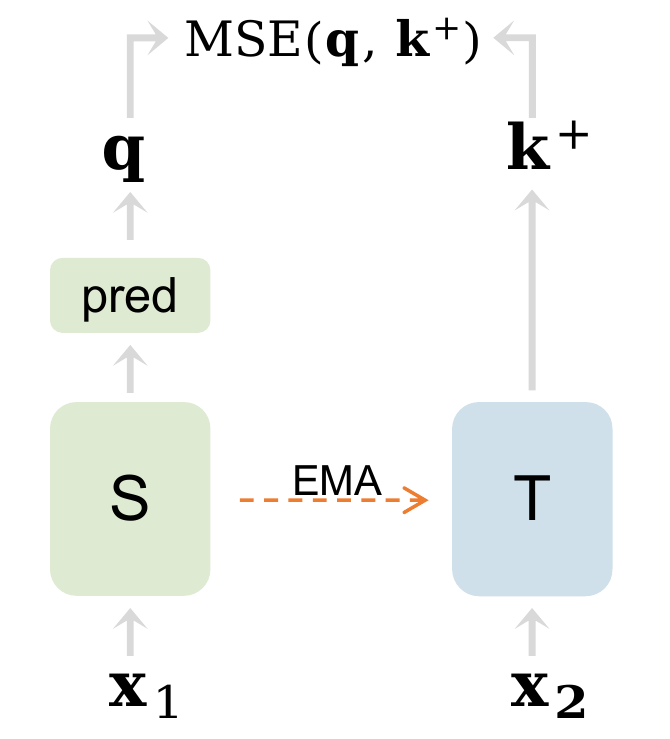}
    \caption{BYOL}
    \label{fig:frame_a}
  \end{subfigure}
  \hfill
  \begin{subfigure}{0.24\linewidth}
    \includegraphics[width=1.0\textwidth]{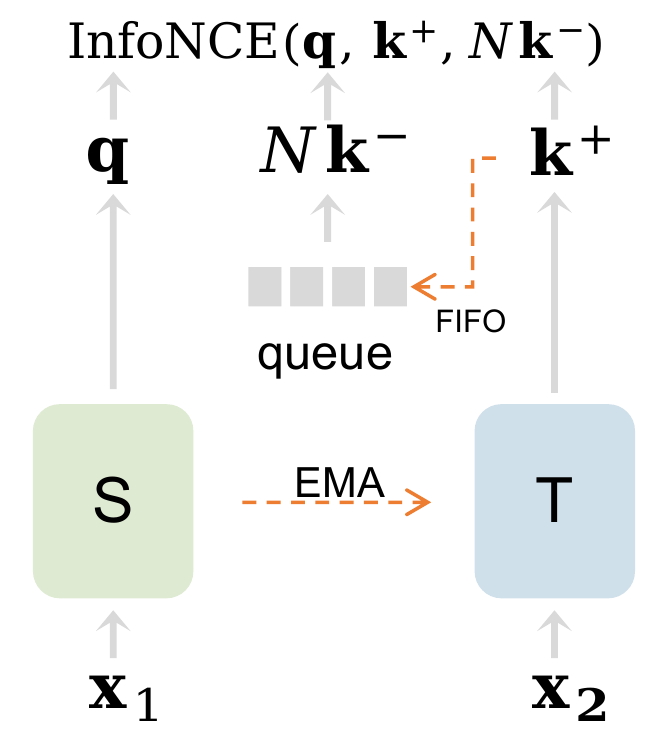}
    \caption{MoCo v2}
    \label{fig:frame_b}
  \end{subfigure}
  \hfill
  \begin{subfigure}{0.24\linewidth}
    \includegraphics[width=1.0\textwidth]{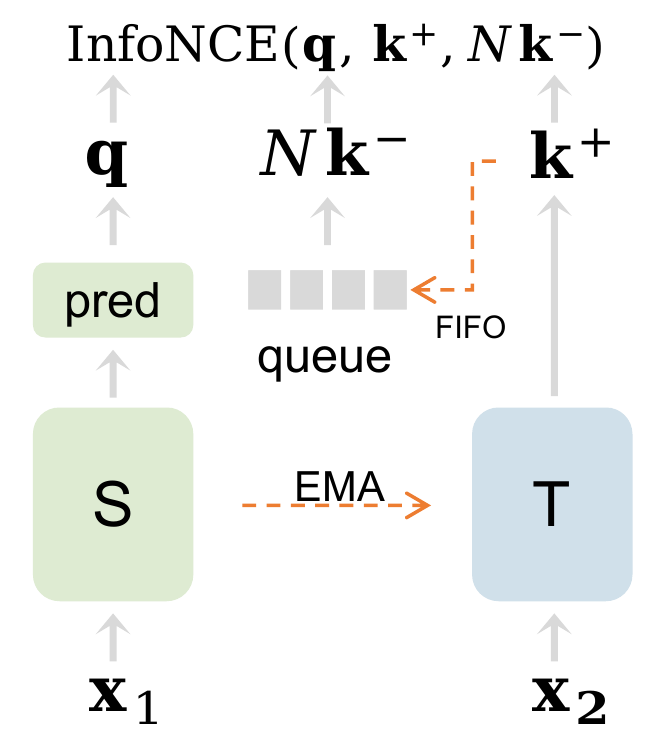}
    \caption{MoCo v2+}
    \label{fig:frame_c}
  \end{subfigure}
  \hfill
  \begin{subfigure}{0.24\linewidth}
    \includegraphics[width=1.0\textwidth]{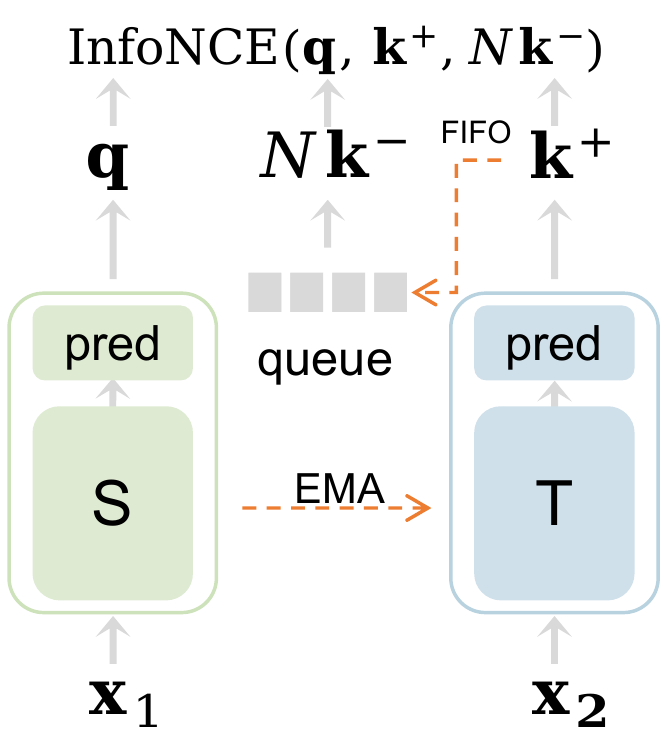}
    \caption{S-MoCo v2+}
    \label{fig:frame_d}
  \end{subfigure}
  \hfill
  \caption{This figure compares the structures of four SSL frameworks discussed in our paper. All of them are siamese network along with the stop-gradient operation and momentum update. For convenient reference, we name the encoder updated by gradients as student encoder, and the encoder with stop-gradient operation as the teacher encoder. They are represented by the capital letter, \textbf{S} and \textbf{T} respectively. Note that the backbones of both student and teacher encoder include a projector that is a non-linear 2-layer MLP (not shown in the picture). \textbf{pred} in the green box represents the predictor (also a non-linear 2-layer MLP). The only difference between \mbox{MoCo v2+} and \mbox{S-MoCo v2+} lies in the existence of teacher encoder's predictor}
  \label{fig:short}
\end{figure*}

We challenge the opinion arising from previous experimental observations of \cite{byol,simsiam} that the superiority of linear evaluation is unique to SSL frameworks without negative sample pairs (e.g., BYOL \cite{byol}, SimSiam \cite{simsiam}). We ablate the differences in model configurations between \mbox{MoCo v2} and BYOL, including network architecture, rule of momentum update, and symmetry of training loss. The differences are iteratively removed based on \mbox{MoCo v2}. Without searching pre-training hyper-parameters, the linear accuracy of \mbox{MoCo v2} on ImageNet consistently benefits from the sophisticated model configurations (72.0\% top-1 accuracy for 200-epoch pre-training). On top of this, we reformulate \mbox{MoCo v2} into a more effective version as shown in \cref{fig:frame_c} (\mbox{MoCo v2+} for short). Moreover, when pre-training with more complex data augmentations, \mbox{MoCo v2+} receives further improvement (72.4\%).  Our study suggests that the sophisticated design of model configurations affects a lot on the pretext task's performance.

Second, we try to uncover the mystery of BYOL’s poor transferability. It seems that practitioners struggle to fully unleash the potential of BYOL even with heavy computation to search fine-tuning learning rates \cite{simsiam}. We tackle this issue by investigating the optimization strategy (e.g., optimizer, learning rate, etc.) of pre-training and fine-tuning. By delving into the original implementation of BYOL and its LARS optimizer \cite{lars}, we find the distribution of LARS-trained representations is different from that of SGD-trained representations. The currently used fine-tuning optimization strategy best selected for SGD-trained features is not suitable for LARS-trained features. We therefore can conclude that the mismatched optimizer choices (LARS for pre-training and SGD for fine-tuning) cause the sub-optimal performances of BYOL. Obviously, using matched optimizers or searching optimization hyper-parameters for fine-tuning can circumvent this issue. In this paper, we also propose one simple yet effective technique NormRescale to solve this problem. NormRescale rescales the weight norm of LARS-trained model by the SGD-trained counterpart. NormRescale works well across many downstream tasks and significantly outperforms the baseline, which proves its capability to recover BYOL's transferability.

Thus far, a fair benchmark has been established. We can make a robust argument that it is not the frameworks but the training details that determine the characteristics of learned representations. 

Thanks to the unified training details, we are able to quest for the experimental impact of the asymmetric network structure. Previous work \cite{byol,simsiam,mocov3} has verified the effectiveness of asymmetric network structure for linear classification. The influences on transfer learning are yet to be examined. To this goal, we symmetrize the network structure of \mbox{MoCo v2+}, which gives us the \mbox{Symmetric MoCo v2+} (abbreviated as \mbox{S-MoCo v2+}, the structure can be seen in \cref{fig:frame_d}). Based on the comparison among \mbox{MoCo v2+}, \mbox{S-MoCo v2+}, and BYOL, our findings are threefold: (i) asymmetric network structure leads to better adaptation on the pre-training datasets but does not mean higher transferability; (ii) the performances of long-tailed classification datasets are more outstanding for contrastive methods, and will be further improved by the symmetric network structure; (iii) contrary to the claim in \cite{byol,barlow_twins}, contrastive methods with or without symmetry of network structure are not more susceptible to data augmentations than BYOL.

Compared to the current literature, our findings are surprising and challenge existing understanding of self-supervised learning. The extensive experiments convey a main idea that \emph{training details determine the characteristics of learned representations}. As long as we align the model configurations, combination of data augmentations and optimization strategy of \mbox{MoCo v2} and BYOL, they show similar performances in linear evaluation and transferring to other downstream tasks. We hope the fair benchmark and our observations will motivate future research.

\section{Related Work}

\noindent\textbf{Augmentation-invariant representation learning.} There have been a great deal of pretext task \cite{exemplar_cnn,context_prediction,rotation,jigsaw,inpainting,cpc,inst_disc,cmc,deep_cluster,moco,pirl,simclr,byol,swav,dcl,propagate_yourself,barlow_twins,whitening,dino,unigrad,adco} proposed in self-supervised learning. Amongst them, augmentation-invariant representation learning shines brightly. The core idea of augmentation-invariant representation learning is to attract different augmented views of the same image as closely as possible. Many research branches are derived from the creative endeavor of the community. Contrastive methods \cite{cpc,cmc,moco,simclr,dcl,propagate_yourself,adco} follow the idea proposed in \cite{contrastive_learning} to pull together the positive sample pairs and push away the negative sample pairs. BYOL \cite{byol} and SimSiam \cite{simsiam} directly minimize the distance of positive sample pairs, along with an asymmetric siamese network. W-MSE \cite{whitening} attracts the positive pairs based on the whitening features. SwAV \cite{swav} first performs online clustering and then classification according to the clustering label generated by its positive sample. DINO \cite{dino} optimizes the distribution distances of positive sample pairs along with the ``centering'' and ``sharpening'' operations. BarlowTwins \cite{barlow_twins} maximizes the correlation of positive sample pairs and decorrelates the features of different images. Research on augmentation-invariant representation learning has sprung up, which also illustrates the advantages of augmentation-invariant representation learning as a pretext task for self-supervised learning. \\

\noindent\textbf{Impact of training details.} Discussion about the impact of training details on representation quality is not new to the community. Previous work has explored which factors enable performance promotion for their algorithms. For example, in order to boost the accuracy of linear evaluation, MoCo \cite{moco,mocov2,mocov3} and SimCLR \cite{simclr,simclr_v2} series search for optimization hyper-parameters (e.g., learning rate, learning rate decay schedule, batch size, etc.), the combination of data augmentations, and the number of negative samples. BYOL \cite{byol,byol_wo_statistics} and SimSiam \cite{simsiam} ablates the coefficients of momentum update and the choice of batch normalization. Due to the lack of a fair benchmark, the successes of these methods seem to be binding together with their unique framework. We are not aware of whether future work can learn from their successes. 

Other work like \cite{id_goodfor_transfer} makes a contribution to better understanding the transfer performance of instance discrimination. But the scope of their study is narrowed down to \mbox{MoCo v2}. SimSiam \cite{simsiam} pays more attention to what the optimization problem for frameworks without using negative samples is. \cite{ssl_transfer} focuses on the fine-tuning results based on frozen pre-trained weights. Unlike them, we provide extensive experiments based on \mbox{MoCo v2} and BYOL that are pre-trained given various training details. The standard evaluation protocol includes linear evaluation on pre-training datasets and transfer performance of some typical downstream tasks. Our goal is to build the first fair benchmark to compare \mbox{MoCo v2} and BYOL, two influential frameworks in augmentation-invariant representation learning.\\

\noindent\textbf{Asymmetric network structure.} The notorious problem of augmentation-invariant representation learning is that without careful design, all input images are mapped to a constant vector. The solution of contrastive frameworks is simple and intuitive---repulsing the negative sample pairs. Likewise, feature decorrelation methods \cite{whitening,dino,feature_decorrelation} separate the features according to the specific rules to avoid the collapse. BYOL \cite{byol} and SimSiam \cite{simsiam} rely on the asymmetric network structure and the stop-gradient operation. 
To study the optimization problem based on asymmetric network structure, SimSiam ablates many hyper-parameters of pre-training. Later work \cite{tian2021understanding} concentrates on the theoretical influence of the asymmetric network structures. 

It should be noted that the focus of this work is not on advancing the development of SSL by proposing a new algorithm. On the opposite, we aim to present a fair and comprehensive investigation based on existing algorithms to gain better understanding.

\section{Experimental Setup}

\subsection{Framework}
\label{sec:review_moco_and_byol}

In this section, we briefly review two well-known frameworks of augmentation-invariant representation learning: \mbox{MoCo v2} \cite{mocov2} and BYOL \cite{byol}. Both of them adopt the design of teacher-student siamese network with momentum update rule \cite{mean_teacher}, where the teacher encoder is updated by the exponential moving average of the student encoder. This unity of network structure is convenient for us to perform controlled experiments. It is worth noting that other self-supervised learning frameworks pre-training with different pretext tasks are beyond the scope of our paper. \\
 
\noindent\textbf{MoCo v2.} By optimizing the contrastive loss \cite{contrastive_learning}, \mbox{MoCo v2} learns to pull the features of positive sample pairs (different augmented views of the same image) together and to push the features of negative sample pairs (different images) away. Different from other contrastive frameworks \cite{cpc,inst_disc,cmc,pirl,simclr}, \mbox{MoCo v2} designs a memory queue (first-in, first-out) to store features computed in previous training iterations. Meanwhile, the rule of momentum update helps maintain the feature consistency. In practice, a batch of input images will be independently transformed twice, resulting in a batch of positive sample pairs. The teacher-student siamese network then encodes them as features respectively. The mini-batch contrastive loss is described as follow:

\begin{equation}
    \label{eq:contrastive_loss}
    L = -\frac{1}{N}\sum_{\mathbf{q}}\log\left(\frac{\exp(\mathbf{q}^{\mathsf{T}}\mathbf{k^{+}}/\tau)}{\exp(\mathbf{q}^{\mathsf{T}}\mathbf{k^+}/\tau) + \sum_{\mathbf{k^{-}}}\exp(\mathbf{q}^{\mathsf{T}}\mathbf{k^{-}}/\tau)}\right)
\end{equation}

\noindent $\mathbf{q}$ and $\mathbf{k^+}$ stand for the student feature and teacher feature that are encoded from the positive sample pair by the siamese network respectively. $\mathbf{k^-}$ is the negative feature stored in the memory queue. $N$ is the batch size and $\tau$ is the temperature (for the following experiments of our paper, we use $0.2$ by default). After back-propagating the contrastive loss, all the teacher features $\{\mathbf{k^+}$\} are enqueued and the ``oldest'' of the memory queue features are dequeued. \\

\noindent\textbf{BYOL.} Similar to contrastive methods, BYOL learns to attract the positive sample pairs as close as possible in feature space without regularization of negative sample pairs. Previous work \cite{byol,simsiam} have stated the asymmetric structure of siamese network and the stop-gradient operation (no gradient will flow to the teacher encoder) are critical to avoiding trivial solution in BYOL. The asymmetric structure refers to that the student branch of siamese network is followed by a predictor (a non-linear two-layer MLP), yielding the asymmetry between student and teacher branches. The mini-batch training loss of BYOL is symmetric:

\begin{equation}
    \label{eq:byol_loss}
    L = \frac{1}{N}(\sum_{\mathbf{q_1}}\|\mathbf{q_1} - \mathbf{k_{1}}\|^2 + \sum_{\mathbf{q_2}}\|\mathbf{q_2} - \mathbf{k_{2}}\|^2)
\end{equation}

\noindent The samples from a positive pair are mapped to $\mathbf{q_1}$ and $\mathbf{q_2}$ by the student encoder, and mapped to $\mathbf{k_1}$ and $\mathbf{k_2}$ by the teacher encoder. $\|\cdot\|$ is the Euclidean distance. 
% In the original implementation of BYOL, there are two combinations of data augmentations which produce similar results, symmetric and asymmetric versions. In this work, we only consider the symmetric one. We provide more details in Supplementary Materials.

\subsection{Pre-training and Evaluation}
\label{sec:pretrain_eval}

In this section, we provide the required information on pre-training and fine-tuning for our experiments. The backbone of siamese network is ResNet-50 \cite{resnet}, and the pre-training dataset is ImageNet \cite{imagenet}. The details about data augmentations can be found in Supplementary Materials.\\

\noindent\textbf{Pre-training.} To re-implement \mbox{MoCo v2} efficiently, we make the following adjustments: increasing the training batch size to 1024, linearly scaling up the learning rate to 0.12 according to \cite{linear_rule}, and introducing a 10-epoch linear warm-up schedule before the decay of learning rate. Note that these modifications do not change the performance of \mbox{MoCo v2}. To reproduce BYOL, we faithfully follow the training settings in \cite{byol}. There are two combinations of data augmentations mentioned in BYOL. We use the symmetric one for our experiments. In the crossover study of \cref{sec:byol_sgd}, when training \mbox{MoCo v2+} with LARS \cite{lars} optimizer, the training hyper-parameters are copied from the implementation of the original BYOL. Likewise, when training \mbox{BYOL} with SGD optimizer, we adopt the same hyper-parameters used in \mbox{MoCo v2+}.\\

\noindent\textbf{Linear evaluation.} The common practice of linear evaluation is to freeze the backbone and train a linear classifier based on the fixed representations. Here, we provide two settings for the training phase of linear evaluation. For models pre-trained with SGD optimizer, we use SGD optimizer to train for $100$ epochs. The batch size is $256$, and the initial learning rate is $30$ which is decayed by a factor of 10 at the $60$ and $80$-th epoch. For models pre-trained with LARS optimizer, we follow the hyper-parameters adopted in BYOL. We use SGD optimizer with Nesterov to train for 80 epochs. The batch size is $1024$, and the initial learning rate is 0.8 and is decayed to $0$ by the cosine schedule. Both training settings use a momentum of $0.9$ and no weight decay. After training, we report the single-crop classification accuracy on ImageNet validation set. \\

\noindent\textbf{PASCAL VOC object detection.} We transfer the pre-trained models on PASCAL VOC \cite{voc} for object detection. We strictly follow the training details in \cite{moco}, which uses a Faster R-CNN \cite{frcnn} detector with a backbone of ResNet50-C4. It takes 9k iterations to fine-tune on \texttt{trainval2007} set and 24k iterations to fine-tune on \texttt{trainval07+12} set. We report the results evaluated on \texttt{test2007} set. \\

\noindent\textbf{COCO object detection and instance segmentation.} We fine-tune the pre-trained models on COCO \cite{coco} for object detection and instance segmentation. We adopt Mask R-CNN \cite{mrcnn} as the detector with two kinds of backbone, ResNet50-C4 and ResNet50-FPN. For a fair comparison, the training settings are exactly the same used in \cite{moco}. Following the $1\!\times$ optimization setting, it takes 90k iterations to fine-tune on \texttt{train2017} set. Finally, we report the results evaluated on \texttt{val2017} set. \\

\noindent\textbf{CityScapes semantic segmentation.} We train on CityScapes \cite{cityscapes} to evaluate the performance on semantic segmentation. For easy re-implementation, we use DeepLab-v3 architecture \cite{deeplabv3}. The backbone is ResNet50 with a stride of 8. The crop size is $512\!\times\!1024$ for training, and $1024\!\times\!2048$ for testing.  It takes 40k iterations to fine-tune on \texttt{train\_fine} set, and finally we report the results evaluated on \texttt{val} set. \\

\section{Experiments and Analyses}

\subsection{What Matters in Linear Evaluations?}
\label{sec:mocov2+}

In the light of previous work, SSL methods without negative sample pairs (e.g., BYOL \cite{byol}, SimSiam \cite{simsiam}, DINO \cite{dino}) have higher accuracy in linear evaluation, compared to contrastive methods. What on earth hinders contrastive methods like \mbox{MoCo v2} from better adapting to the pre-training dataset to achieve better performance in linear evaluation, the negative sample pairs or other previously ignored factors? 

In this subsection, we seek to answer the question by exploring how to elevate \mbox{MoCo v2} to achieve higher accuracy in linear evaluation. Given this goal, it is desirable to improve linear accuracy under modifications to model configurations. With reference to BYOL, we make the following adjustment on \mbox{MoCo v2}. First, we replace the ShufflingBN with synchronized BN (SyncBN). However, this direct replacement does not bring the expected performance improvement. Hence we insert a BN to the hidden layer of the projector (a non-linear two-layer MLP after the backbone) as BYOL does. Second, we add a predictor (an MLP similar to projector) at the end of student encoder, yielding asymmetry between student encoder and teacher encoder. Third, the coefficient of momentum update no longer stays still, but increases from $0.99$ to $1$ according to a cosine schedule. Forth, we symmetrize the contrastive loss, as has been done in \cite{simclr,byol,swav,simsiam}. For convenient reference, we name this enhanced framework as \mbox{MoCo v2+}, which is an extension of \mbox{MoCo v2}. Last, we train \mbox{MoCo v2+} with more complex data augmentations (introducing solarization to the combination)\footnote{The detailed information about the combination of data augmentations can be found in Supplementary Materials}.

\begin{figure*}[t]
    \centering
    \includegraphics[width=0.5\textwidth]{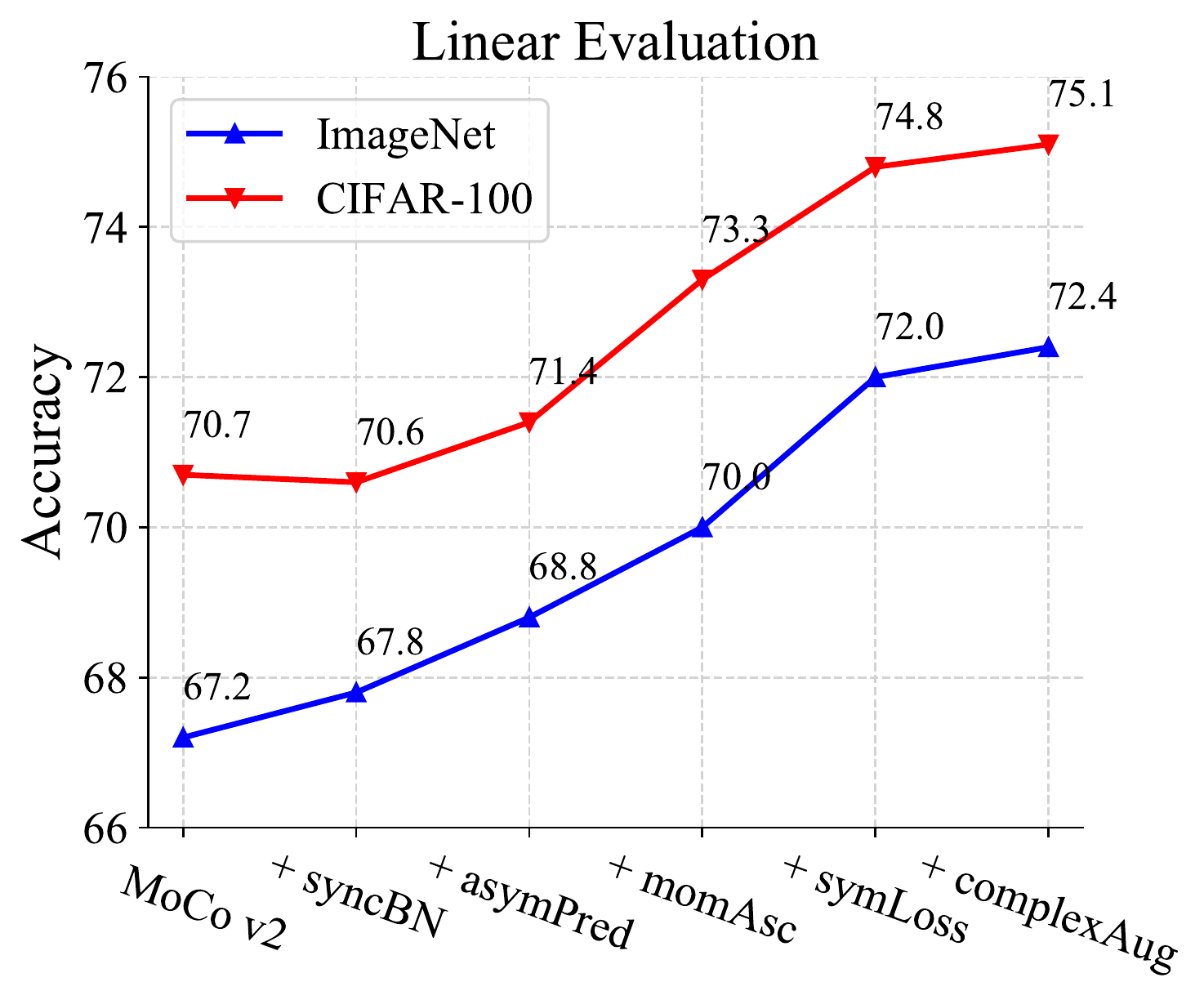}
    \caption{Top-1 accuracy of linear evaluation on ImageNet and CIFAR-100. The x-axis represents the modifications of model configurations. We use \mbox{MoCo v2} as our baseline. All models are trained for 200 epochs. The trend of these two curves indicates that the linear accuracy consistently benefits from the sophisticated model configurations}
    \label{fig:linear_eval}
\end{figure*}

\begin{table*}[t]
    \centering
    \caption{The results of transfer learning on detection and segmentation tasks. All models are trained for 200 epochs. The best results are marked as bold}
    \label{tab:model_config}
    \scalebox{0.9}{
    \begin{tabular}{l|c c|c c c c|c}
    \multirow{2}{*}{} & VOC07 & VOC07+12 & \multicolumn{4}{c|}{COCO} & CityScapes  \\
    \cline{2-8}
    & $\text{AP}_{50}$ & $\text{AP}_{50}$ & $\text{AP}_{\text{box}}^{\text{C4}}$ &
    $\text{AP}_{\text{seg}}^{\text{C4}}$ & $\text{AP}_{\text{box}}^{\text{FPN}}$ &  $\text{AP}_{\text{seg}}^{\text{FPN}}$ & mIoU \\
    \toprule
    \mbox{MoCo v2} & 76.5 & 82.2 & 38.8 & 34.0 & 39.5 & 35.8 & 77.4 \\
    + SyncBN & 76.7 & 82.0 & 38.6 & 33.7 & 39.7 & 35.8 & 76.9 \\
    + Asymmetric Predictor & 76.7 & 82.1 & 39.0 & 34.1 & 39.8 & 35.9 & 77.3 \\
    + Momentum Ascending & 77.0 & 82.3 & 39.0 & 34.3 & 39.7 & 35.8 & 77.4 \\
    + Symmetric Loss (MoCo v2+) & 77.1 & \textbf{82.7} & \textbf{39.4} & \textbf{34.5} & 40.3 & 36.5 & \textbf{77.6} \\
    + More Complex Augmentations & \textbf{77.3} & \textbf{82.7} & 39.2 & 34.4 & \textbf{40.4} & \textbf{36.7} & \textbf{77.6} \\
    \midrule
    BYOL& 71.7 & 79.1 & 35.3 & 31.1 & 40.8 & 36.9 & 76.4  \\
    \end{tabular}}
\end{table*}

The results of linear evaluation on ImageNet are in \cref{fig:linear_eval}. Surprisingly, the linear accuracy consistently benefits from the modifications even without searching hyper-parameters. When training with more complex augmentations, \mbox{MoCo v2+} finally catches up to BYOL in terms of linear accuracy (72.4\% top-1 accuracy). Among these changes, the symmetrization of contrastive loss brings the most obvious improvement (2.0\% accuracy increment). To validate the effectiveness of representations with high linear accuracy on ImageNet, we train a linear classifier on CIFAR-100 \cite{cifar}. The accuracy curve is in \cref{fig:linear_eval}. Similarly, we observe distinct promotions for linear evaluation on CIFAR-100 compared to baseline.

\cref{{tab:model_config}} presents the transfer performances on downstream tasks. The promotions (about 0.5\% improvement) in transfer learning are not as obvious as in linear evaluation. One possible explanation for this contradiction is that better adaptation to pre-training dataset is more helpful to those datasets whose distribution are similar to the pre-training dataset. As we can see in \cref{fig:linear_eval}, the trends of two curves in \cref{fig:linear_eval} are accordant. In a nutshell, \emph{sophisticated design of model configurations affects a lot on the pretext task's performance}.

\subsection{How to Improve Transfer Performances?}
\label{sec:byol_sgd}

Despite BYOL being one of the significant frameworks in SSL, its capability on typical downstream tasks like object detection on VOC \cite{voc} and COCO \cite{coco} have not received enough attention. From one of only a few studies concerning this problem, we find \mbox{MoCo v2} outperforms BYOL on VOC and COCO detection and instance segmentation \cite{simsiam}. \cref{tab:model_config} also reflects this issue. We notice these comparisons are based on misaligned optimization strategies. Specifically, BYOL utilizes LARS optimizer~\cite{lars} to train with large batch size, while \mbox{MoCo v2+} uses SGD optimizer. In this case, it remains an open question whether BYOL has innately poor transferability on those challenging downstream tasks given the same optimization strategy. 

In this subsection, we investigate how to improve transfer performances of BYOL from the perspective of optimization. To understand how optimization strategy influences the transferability of learned representations, we provide a crossover study of SGD and LARS optimizers for pre-training. The results of downstream tasks are in \cref{tab:optimizer}. Both frameworks are less competitive on most downstream tasks when pre-trained with LARS. It seems that the poor results may originate from the pre-training optimization strategy. There is one exception, though, that LARS-trained models show comparable or even better results for the downstream tasks adopting ResNet50-FPN as backbone. We, therefore, infer that the LARS optimizer does not compromise the quality of learned representations. 

\begin{table*}[t]
    \centering
    \caption{The results of crossover study involving SGD and LARS optimizer. All models are trained for 200 epochs. The best results are marked bold}
    \scalebox{0.9}{
    \begin{tabular}{l|c|c|c c|c c c c|c}
     & \multirow{2}{*}{Optimizer} & ImageNet & VOC07 & VOC07+12 & \multicolumn{4}{c|}{COCO} & CityScapes\\
    \cline{3-10}
    & & Acc & $\text{AP}_{50}$ & $\text{AP}_{50}$ & $\text{AP}_{\text{box}}^{\text{C4}}$ &
    $\text{AP}_{\text{seg}}^{\text{C4}}$ & $\text{AP}_{\text{box}}^{\text{FPN}}$ &  $\text{AP}_{\text{seg}}^{\text{FPN}}$ & mIoU\\
    \toprule
    % MoCo v2 (BYOL-Aug) & 66.8 & 81.8 & & 38.6 & 33.8 & 39.8 & 35.9 & 78.4\\
    \multirow{2}{*}{MoCo v2+} & SGD & 72.0 & \textbf{77.1} & \textbf{82.7} & \textbf{39.4} & \textbf{34.5} & 40.3 & 36.5 & \textbf{77.7}\\ 
    & LARS & \textbf{72.5} & 62.9 & 74.4 & 32.1 & 28.9 & 40.0 & 36.2 & 73.2 \\
    \midrule
    \multirow{2}{*}{BYOL} & SGD & 72.1 & 76.2 & 82.4 & 38.8 & 33.9 & 39.9 & 36.1 & 77.5 \\
     & LARS & 72.4 & 71.7 & 79.1 & 35.3 & 31.1 & \textbf{40.8} & \textbf{36.9} & 75.2 \\
    \end{tabular}}
    \label{tab:optimizer}
\end{table*}

\begin{figure*}[t]
  \centering
  
  \begin{subfigure}{0.54\linewidth}
    \includegraphics[width=1.0\textwidth]{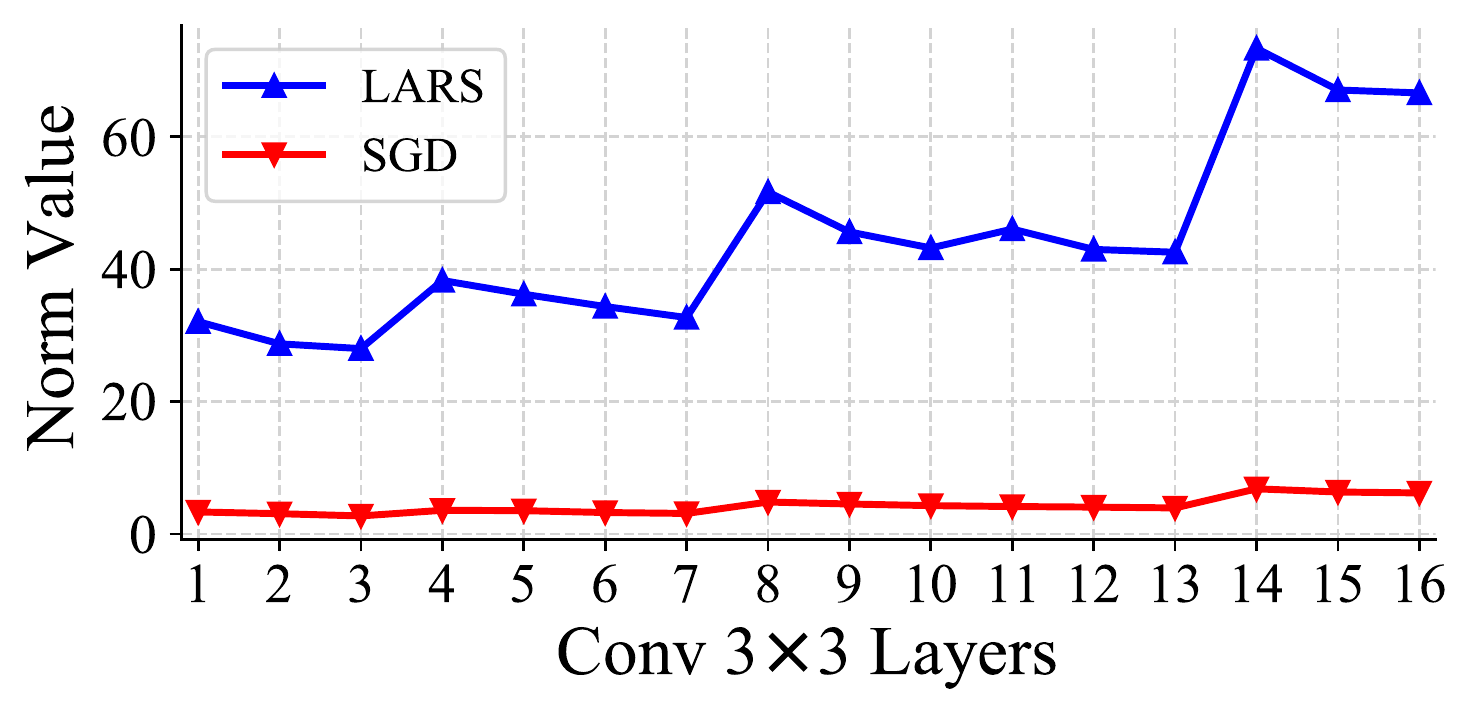}
    \caption{Weight norm}
    \label{fig:weight_norm}
  \end{subfigure}
  \hfill
  \begin{subfigure}{0.44\linewidth}
    \includegraphics[width=1.0\textwidth]{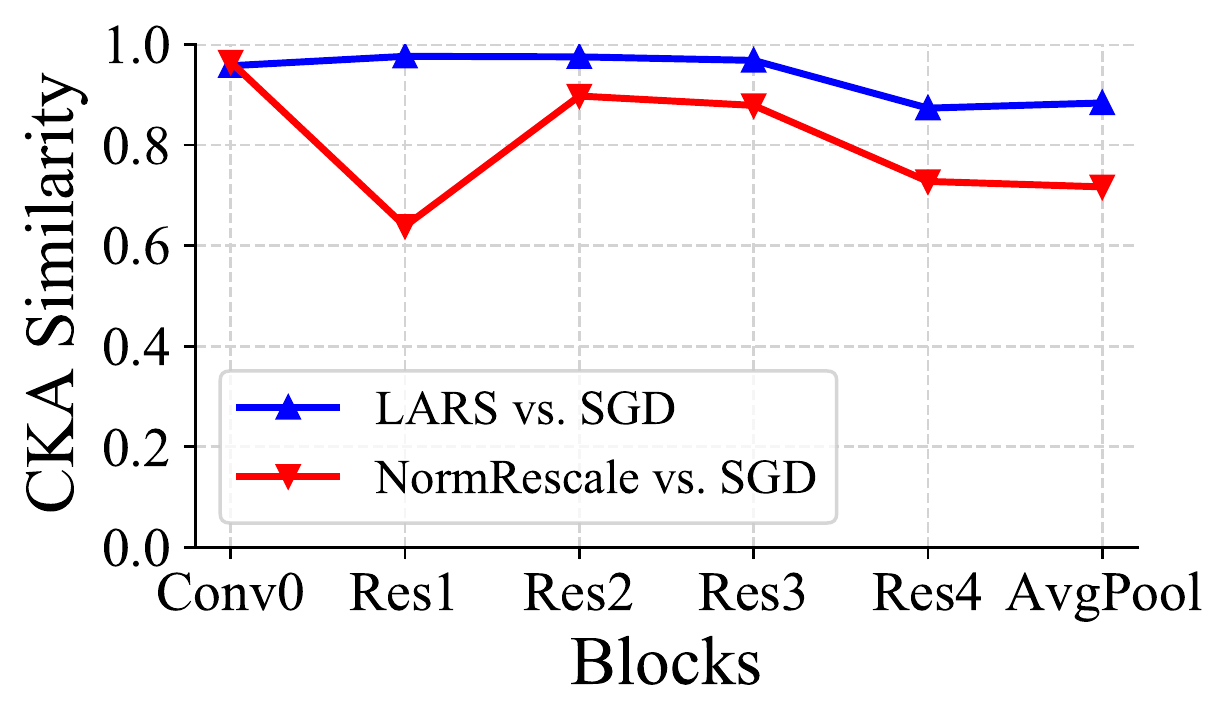}
    \caption{CKA similarity}
    \label{fig:cka}
  \end{subfigure}
  \hfill
  
  \caption{(\textbf{a}): Weight norms of all \mbox{conv$3\times3$} layers from LARS-trained and SGD-trained models. (\textbf{b}): CKA similarities of LARS-trained and SGD-trained representations across all stages of ResNet-50. Best viewed in color} 
  \label{fig:optimization}
\end{figure*}

By examining the implementation details of BYOL, we find that, unlike SGD, the LARS optimizer does not impose L2-regularization on the parameters of batch normalization layers. As training goes on, the weight norm becomes larger. We can see the clear contrast in \cref{fig:weight_norm} that the weight norms of the LARS-trained model are significantly larger. In other words, the distribution of learned representations is different from those trained by SGD. Fine-tuning LARS-trained models with the hyper-parameters best suited for SGD-trained models naturally yields sub-optimal performances. Thus, we can conclude that \emph{mismatched optimizer used in pre-training and fine-tuning is the reason for performance degeneration in BYOL, but not the framework itself.} 

\cref{tab:optimizer} points out a solution to circumvent this issue---using SGD optimizer for pre-training. This solution, however, is not universally effective, since it does not apply to large batch size training where LARS is more popularly used. To alleviate this issue, \cite{simsiam} searches learning rates for fine-tuning LARS-trained models, inevitably inducing heavy computation. Next, we describe two findings that lead us to a flexible approach.

First, we utilize the CKA similarity \cite{cka} to measure how similar the representations learned by LARS and SGD are. The blue line of \cref{fig:cka} indicates these representations are sufficiently similar although they follow different distributions. Second, as described in \cite{moco}, the features for the region proposal are normalized by the newly initialized BN in ResNet50-FPN, while not in ResNet50-C4. We argue the rescale operation in newly initialized BN helps LARS-trained models to adapt to optimization of fine-tuning driven by SGD. Motivated by the analyses above, we present a simple yet effective technique, NormRescale, to address this issue. Assume we have a well-trained model that is pre-trained by SGD\footnote{In default, we choose the 200-epoch SGD-trained BYOL.}. For any weight of the LARS-trained model, we rescale its norm as follows:

\begin{equation}
    {\mathbf{w}^{\ast}} = \lVert \mathbf{w}_{\text{S}}\rVert \cdot \frac{\mathbf{w}_{\text{L}}}{\lVert \mathbf{w}_{\text{L}}\rVert },
\end{equation}

\noindent where $\mathbf{w}_{\text{L}}$ is the weight vector of LARS-trained model, and $\mathbf{w}_{\text{S}}$ is the corresponding weight vector of SGD-trained model. $\lVert\cdot\rVert$ stands for 2-norm. We skip the procedure of hyper-parameters searching and fine-tune the processed weight $\mathbf{w}^{\ast}$ on downstream tasks. 

In \cref{tab:renorm}, we compare the transfer performances of \mbox{BYOL} under different implementations. The results of NormRescale are about the same as that of \mbox{BYOL-SGD} and significantly better than that of the vanilla implementation (\mbox{BYOL-LARS}). Moreover, it also shows superior performances on most metrics for detection and segmentation against the reproduction of \cite{simsiam}. The comparisons confirm NormRescale can effectively recover the transferability of the LARS-trained model. We also plot the CKA similarities between the representations of BYOL-SGD and NormRescale in \cref{fig:cka} (red line). It can be seen that NormRescale retains the characteristics of LARS-trained representations. Apart from using \mbox{BYOL-SGD} as the anchor weight, we also explore other anchor choices for NormRescale. The detailed comparison can be found in Supplementary Materials.

These results suggest that optimization strategy is the key to the transferability of BYOL. For efficient comparison, we adopt SGD as our default optimizer for all the below experiments. Without confusing references, we continue to use BYOL to stand for SGD-trained BYOL.

\begin{table}[t]
    \begin{center}
    \caption{The downstream performances of BYOL under various implementations. 
    ``NR'' stands for NormRescale. All models are trained for 200 epochs. The best results are marked bold}
    \label{tab:renorm}
    \scalebox{0.95}{
    \begin{tabular}{l|c c c|c c c|c c c|c c c}
     & \multicolumn{3}{c|}{VOC07} & \multicolumn{3}{c|}{VOC07+12} & \multicolumn{3}{c|}{COCO detection} & \multicolumn{3}{c}{COCO instance seg.} \\
     & $\text{AP}_{50}$ & AP & $\text{AP}_{75}$ &$\text{AP}_{50}$ &  AP & $\text{AP}_{75}$ & $\text{AP}_{50}$  & AP & $\text{AP}_{75}$ & $\text{AP}_{50}^{M}$ & $\text{AP}^{M}$ & $\text{AP}_{75}^{M}$ \\
    \toprule
    BYOL-SGD & 76.2  & \textbf{48.1}  & \textbf{52.9}  & \textbf{82.4}  & 56.5  & \textbf{63.6}  & 58.5  & 38.8  & 42.1  & 55.2  & 34.0  & 36.2  \\
    BYOL-LARS & 71.7  & 38.8  & 37.0  & 79.1  & 48.7  & 51.7  & 56.2  & 35.3  & 37.5  & 52.3  & 31.1  & 32.2  \\
    BYOL in \cite{simsiam} & \textbf{77.1}  & 47.0  & 49.9  & 81.4  & 55.3  & 61.1  & 57.8  & 37.9  & 40.9  & 54.3  & 33.2  & 35.0  \\
    BYOL-NR & 76.6  & \textbf{48.1}  & 51.6  & 82.1  & \textbf{56.7}  & 62.9  & \textbf{59.3}  & \textbf{39.3}  & \textbf{42.6}  & \textbf{56.0}  & \textbf{34.5}  & \textbf{36.7}  \\
    \end{tabular}}
    \end{center}
\end{table}

Thus far, we have presented the first fair benchmark to compare two important frameworks of SSL, namely \mbox{MoCo} and BYOL. Our extensive experiments show that the performances of linear evaluation and transfer learning are similar in \mbox{MoCo v2+} and BYOL given the aligned training details, leading to an authentic argument that \emph{the training details determine the characteristics of learned representations}.

\subsection{What Is the Impact of Asymmetric Network Structure?}
\label{sec:asymmetric_network_impact}

The asymmetric network structure first proposed in BYOL plays an important role in avoiding model collapse for augmentation-invariant representation learning. There are follow-up studies on the asymmetric structure that are mainly about the theoretical understanding \cite{tian2021understanding,simsiam} and the effectiveness for linear classification \cite{simsiam,mocov3}. Here, we explore the experimental impact of asymmetric structure in transfer learning and its comparison to symmetric one. We symmetrize the network structure of \mbox{MoCo v2+} by adding an extra predictor to the teacher encoder. We call it Symmetric \mbox{MoCo v2+} (abbreviated as \mbox{S-MoCo v2+}). We refer to \cref{fig:frame_d} for visual description. In this subsection, the experiments are mainly about the following three parts.\\

\noindent\textbf{Standard evaluation tasks.}  We first provide a direct comparison amongst \mbox{MoCo v2+}, \mbox{S-MoCo v2+}, and BYOL on the typical datasets. The results of linear evaluation and transfer learning are listed in \cref{tab:asym_imgnet_downstream}. As shown in the third column (ImageNet Acc), \mbox{S-MoCo v2} is inferior in linear evaluation, indicating that models with asymmetric structure may better fit pretext tasks. The performances of transfer learning, in contrast, are similarly good on many downstream tasks. The biggest gap between the best and the worst is within 0.4. We conclude that the transferability for these regular downstream tasks may be neutral to the symmetry of network structure. \\

\begin{table*}[t]
    \centering
    \caption{Results on ImageNet and downstream tasks of BYOL, \mbox{MoCo v2+}, \mbox{S-MoCo v2+}. All models are trained for 200 epochs. The best results are marked bold}
    \label{tab:asym_imgnet_downstream}
    \scalebox{0.87}{
    \begin{tabular}{l|c|c|c c|c c c c|c}
     & \multirow{2}{*}{Asymmetry} & ImageNet & VOC07 & VOC07+12 & \multicolumn{4}{c|}{COCO} & CityScapes\\
    \cline{3-10}
    & & Acc & $\text{AP}_{50}$ & $\text{AP}_{50}$ & $\text{AP}_{\text{box}}^{\text{C4}}$ &
    $\text{AP}_{\text{seg}}^{\text{C4}}$ & $\text{AP}_{\text{box}}^{\text{FPN}}$ &  $\text{AP}_{\text{seg}}^{\text{FPN}}$ & mIoU\\
    \toprule
    % MoCo v2 (BYOL-Aug) & 66.8 & 81.8 & & 38.6 & 33.8 & 39.8 & 35.9 & 78.4\\
    BYOL & $\checkmark$ & 72.1 & 77.2 & \textbf{82.7} & \textbf{39.3} & \textbf{34.5} & \textbf{40.6} & 36.6 & 77.1 \\
    MoCo v2+ & $\checkmark$ & \textbf{72.4} & \textbf{77.3} & \textbf{82.7} & 39.2 & 34.4 & 40.4 & 36.7 & \textbf{77.6} \\
    S-MoCo v2+ &  & 71.2 & 77.1 & 82.4 & 39.1 & 34.2 & \textbf{40.6} & \textbf{36.7} & 77.2 \\

    \end{tabular}}
\end{table*}

\begin{figure}[t]
    \centering
    \includegraphics[width=1.0\textwidth]{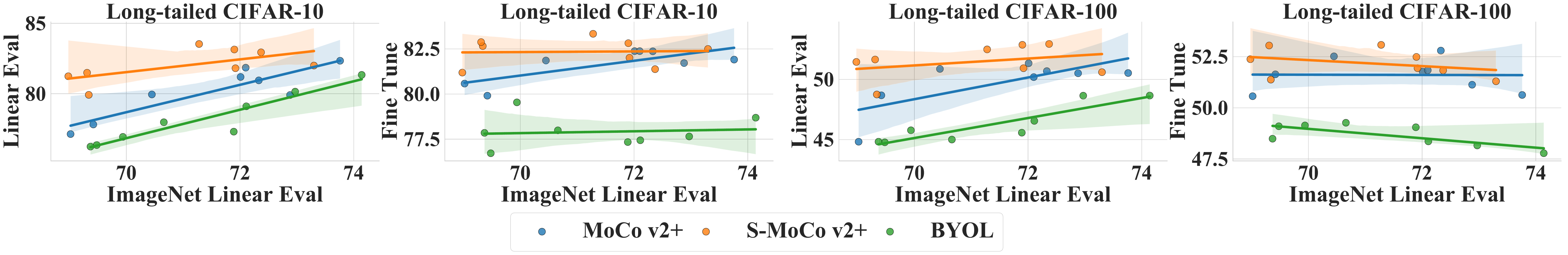}
    \caption{Linear evaluation and fine-tuning results of BYOL, \mbox{MoCo v2+} and \mbox{S-MoCo v2+}.  The regression lines describe the correlation between linear accuracy on ImageNet and linear or fine-tuning accuracy on long-tailed classification datasets, with confidence intervals in shaded areas. Best viewed in color}
    \label{fig:longtail}
\end{figure}

\noindent\textbf{Long-tailed classification task.} We next study the effects on two long-tailed classification tasks (CIFAR-10-LT and CIFAR-100-LT \cite{cifar_lt}). The effectiveness of pre-trained models is measured in two aspects: linear evaluation and fine-tuning. For solid comparisons, we provide models pre-trained with different hyper-parameters (e.g., learning rate, training epochs, etc.). As plotted in \cref{fig:longtail}, the horizontal coordinate for each point represents the pre-trained model's linear accuracy on ImageNet and the vertical coordinate stands for linear or fine-tuning accuracy on long-tailed datasets. Our findings can be summarized as follows:

(i) Model with higher linear accuracy on ImageNet shows better performance on CIFAR-10/100-LT under the linear evaluation protocol. But the situation is different for fine-tuning. We do not see a clear trend between linear accuracy on ImageNet and fine-tuning accuracy. 

(ii) In all four sub-figures, we can observe a clear ranking result that \mbox{S-MoCo v2+} has the best effect, followed by \mbox{MoCo v2+}, and finally BYOL. The superiority of contrastive methods in linear evaluation and fine-tuning implies that the regularization imposed by pushing away negative sample pairs which renders a more uniform representation space is conducive to long-tailed classification. Besides, we point out the strength of symmetric network structure, as it provides the best performances of \mbox{S-MoCo v2+} (see \cref{fig:longtail}).\\

\begin{figure*}[t]
    \centering
    \includegraphics[width=1.0\textwidth]{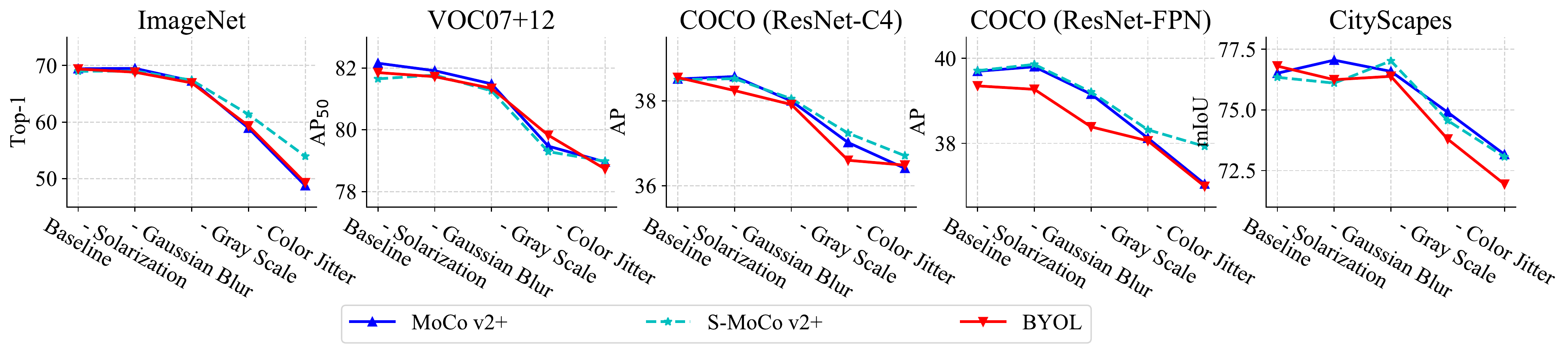}
    \caption{The results of linear evaluation and downstream tasks under different combinations of data augmentations. Best viewed in color}
    \label{fig:remove_augmentation}
\end{figure*}

\noindent\textbf{Data augmentations.} We make an attempt to investigate the sensitivity of contrastive methods and BYOL to data augmentations, which has been discussed in \cite{byol,barlow_twins}. The conclusions about this problem are consistent in their work---contrastive methods are more sensitive to the variation of data augmentations. Following our above analyses, we are sceptical about the validity of this conclusion where a fair benchmark is absent. To get a clear picture of it, we ablate data augmentations based on \mbox{MoCo v2+}, \mbox{S-MoCo v2+}, and BYOL. The baseline combination of data augmentations includes random cropping and resizing to $224\!\times\!224$, horizontal flipping, color jittering, gray scale converting, Gaussian blur, and solarization. The specific parameters of augmentations can be found in Supplementary Materials. Likewise, we iteratively remove the data augmentations involving color transformations. The order goes solarization, Gaussian blur, gray scale converting, and color jittering. After removing all color transformations, % the combination is the same as used in supervised training. The results are depicted in \cref{fig:remove_augmentation}.
the combination is the same as used in supervised training. The results are depicted in \cref{fig:remove_augmentation}.

Interestingly, the obvious accuracy gap between contrastive methods and BYOL reported in \cite{byol,barlow_twins} vanishes; instead, we observe similar results of linear accuracy on ImageNet for contrastive methods and BYOL. The comparison of contrastive methods (\mbox{MoCo v2+} vs.\ \mbox{S-MoCo v2+}) demonstrates that it is not the asymmetric head that causes these similarities. Therefore, we can present a convincing and empirically verified conclusion that a sufficiently complex combination of data augmentation is equally important for contrastive methods and BYOL. In turn, the consistency of effects between \mbox{MoCo v2+}, \mbox{S-MoCo v2+} and BYOL suggests that ignoring training details can give misperceptions in SSL.

In addition to the linear accuracy on ImageNet, we also report the transfer performances across various downstream tasks. As clearly shown in \cref{fig:remove_augmentation}, similar phenomena can be found in the results of different downstream tasks that transferability is positively correlated to the complexity of data augmentation combinations. Through careful observation, we find that most models meet significant performance degeneration if gray scale converting is cancelled. Strictly speaking, it is biased to believe gray scale converting is so important that SSL methods would face degradation without it. A more likely explanation is that the combination of data augmentations lacks complexity when cancelling out gray scale converting, inducing less competitive representations. 

% \section{Limitations}

% We choose two representative frameworks (BYOL and \mbox{MoCo v2}) as the experimental subjects, and other self-supervised learning methods with different pretext tasks are beyond the scope of our discussion. In practice, we only use ResNet-50 as backbone and ImageNet as the pre-trained dataset, which is the most commonly used in self-supervised learning. Instead of searching for hyper-parameters, we use the same optimization strategy to train all models, and we speculate that the optimization hyper-parameters that we use may not be best suited for all models. However, we mainly focus on the trend variation and the relative improvement or degradation under fair comparison. We can also observe performance promotion compared to the results reported in the original paper. Therefore, we believe our derivations can generalize broadly in self-supervised learning.\\

\section{Conclusion}

In summary, the extensive experiments throughout the paper revolve around the idea that training details determine the characteristics of learned representations in augmentation-invariant representation learning. In the process of verifying the idea, we observe the following:

(i) Sophisticated design of model configurations helps representations better adapt to the pre-training dataset, which in turn improves the linear accuracy on datasets with similar distribution to pre-training dataset.

(ii) What truly prevents BYOL from achieving competitive performances on typical downstream tasks is the mismatched optimization strategy for pre-training and fine-tuning. Using matched optimizers can remedy the performances drop. We also propose one simple yet effective technique to do the same, and it can apply to the situation where using mismatched optimizers is inevitable.

(iii) Asymmetric network structure leads to higher linear accuracy on pre-training dataset, while symmetric one has more competitive results on long-tailed classification tasks. Based on the fair comparisons among \mbox{MoCo v2+}, \mbox{S-MoCo v2+} and BYOL, we confirm that contrastive methods and BYOL are equally sensitive to data augmentations.

We hope the fair benchmark and our observations will shed light on the understanding of \mbox{MoCo v2} and BYOL, and help motivate future research to push forward the frontier of SSL. \\

{\bfseries\noindent Acknowledgements. } This research was supported by National Key R\&D Program of China (No. 2017YFA0700800) and Beijing Academy of Artificial Intelligence (BAAI).

\bibliographystyle{splncs04}
\bibliography{simplified_reference}

\newpage
\appendix
\section{Details for Experimental Setup}

\subsection{Data Augmentations}

In \cref{sec:asymmetric_network_impact}, we ablate the data augmentations in \mbox{MoCo v2+} and BYOL. The combinations of data augmentations we used are the same as the symmetric version proposed in \cite{byol}, including random cropping, resizing, horizontal flipping, color jittering, gray scale converting, Gaussian blurring, and solarization. The probability and parameters of the data augmentations are detailed in \cref{tab:aug_parameters}.

\begin{table}[h]
    \centering
    \caption{The probabilities and parameters of the data augmentations}
    \label{tab:aug_parameters}
    \begin{tabular}{l|c|c}
    Data Augmentations & Probability & Parameters\\
    \toprule
    Random Crop & 1.0 & $\left(0.08, 1.0\right)$\\
    Resize & 1.0 & 224\\
    Horizontal Flip & 0.5 & -\\
    Color Jitter & 0.8 & (0.4, 0.4, 0.2, 0.1)\\
    Gray Scale & 0.2 & -\\
    Gaussian Blur & 0.5 & (0.1, 2.0)\\
    Solarization & 0.2 & 128\\
    \end{tabular}
    
\end{table}

\section{More Experiments}

We provide other interesting experiments in Supplementary Materials.

\subsection{Longer Pre-training}

\begin{figure*}[h]
    \centering
    \includegraphics[width=1.0\textwidth]{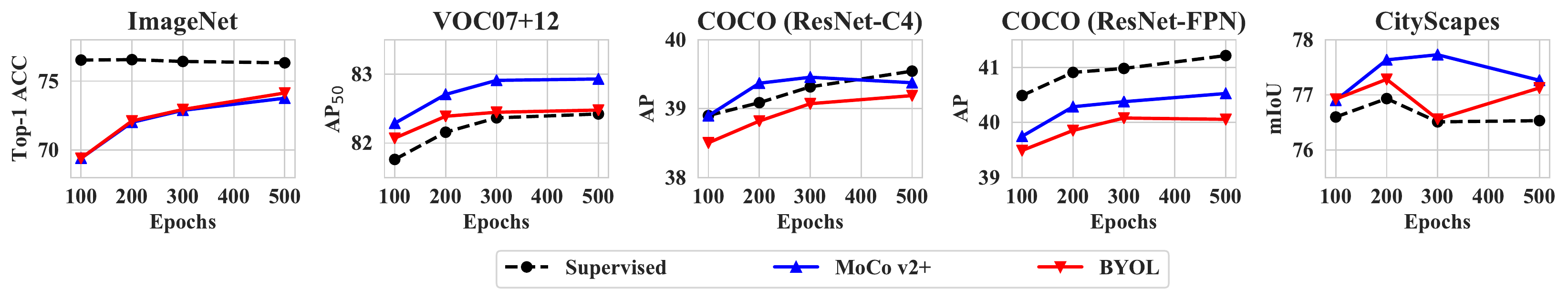}
    \caption{The linear evaluation and transfer learning results of different pre-training methods with various training time}
    \label{fig:longer_training}
\end{figure*}

Here, we benchmark the \mbox{MoCo v2+} and \mbox{BYOL} with longer training. We also introduce the supervised pre-training as a baseline. We start by investigating 12 pre-trained models that vary along two dimensions, training time (i.e., 100, 200, 300, 500 epochs) and pre-training methods (i.e., supervised, \mbox{MoCo v2+}, \mbox{BYOL}). Furthermore, we plot the results of downstream tasks in \cref{fig:longer_training}. For most downstream tasks, it can be seen that all pre-training methods tend to saturation or even degradation with longer training time (e.g., 500 epochs). On the contrary, the linear classification accuracy consistently benefits from longer training. This contradiction unveils the fact that longer training in self-supervised learning helps models better adapt to the pre-training dataset, and does not automatically improve the quality of learned representations.

\begin{figure*}[h]
    \centering
    \includegraphics[width=1.0\textwidth]{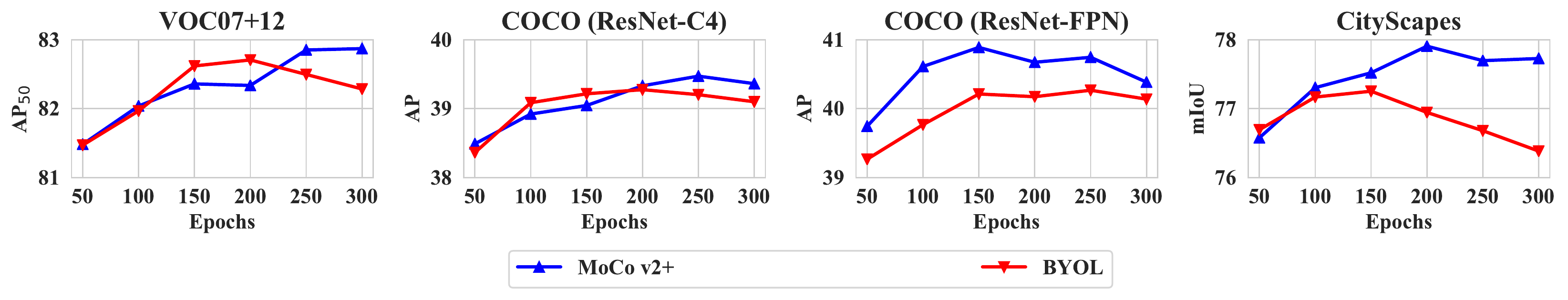}
    \caption{The transfer performances of intermediate checkpoints}
    \label{fig:overfitting}
\end{figure*}

The invalidity of longer training invokes us to examine the overfitting problem in self-supervised learning. We train \mbox{MoCo v2+} and \mbox{BYOL} on ImageNet for 300 epochs. During training, we fetch the intermediate checkpoints for every $50$ epoch, and evaluate them on downstream tasks to see how representations evolve as the optimization proceeds.

The results are plotted in \cref{fig:overfitting}. Without exception, the performances on all downstream tasks meet saturation or even decline after a period of training time, suggesting that overfitting to pretext tasks does happen in self-supervised learning. According to \cite{visualizing_cnn}, lower layers of convolutional neural networks converge rapidly during training. As for self-supervised pre-training, low-level and mid-level representations that are of more importance in transferring to downstream tasks \cite{id_goodfor_transfer} may stop evolving when the learning rate decays to a small value. We, therefore, hypothesize that currently used optimization strategies (e.g., learning rate decays to zero) are the reason for overfitting. A better optimization strategy is waiting to be developed. 

%Hypothetically, the reason for the opposite conclusions is twofold. First, the improvement of longer optimization in their experiments is minor in self-supervised learning. Second, they pay more attention to the AP metric, while we focus on $\text{AP}_{50}$. Performance saturation and degradation of $\text{AP}_{50}$ can also be found in their experiments. We further verify our derivation with the following experiment. We evaluate the 100th checkpoint of \mbox{MoCo v2+} during a 200-epoch training. The AP of COCO detection with the backbone of ResNet-FPN is $40.4$ for the 100th epoch checkpoint of \mbox{MoCo v2+}, which is slightly better than the result of the final model (i.e., $40.3$). Given this counterfactual finding, we expect a better optimization strategy for self-supervised training in future work.

\subsection{Data Variation}

\noindent\textbf{Variation of pre-training data.} Inspired by \cite{scale_benchmark}, we explore the relationship between self-supervised learning methods and the size of pre-training data. We uniformly sample some classes in ImageNet. For any sampled classes, all its training images will be added to the training set. There are five ratios of sampling: $10\%$, $20\%$, $50\%$, $70\%$, $100\%$. To keep training iterations constant, we extend the training epochs adaptively.

\begin{figure*}[t]
    \centering
    \includegraphics[width=1.0\textwidth]{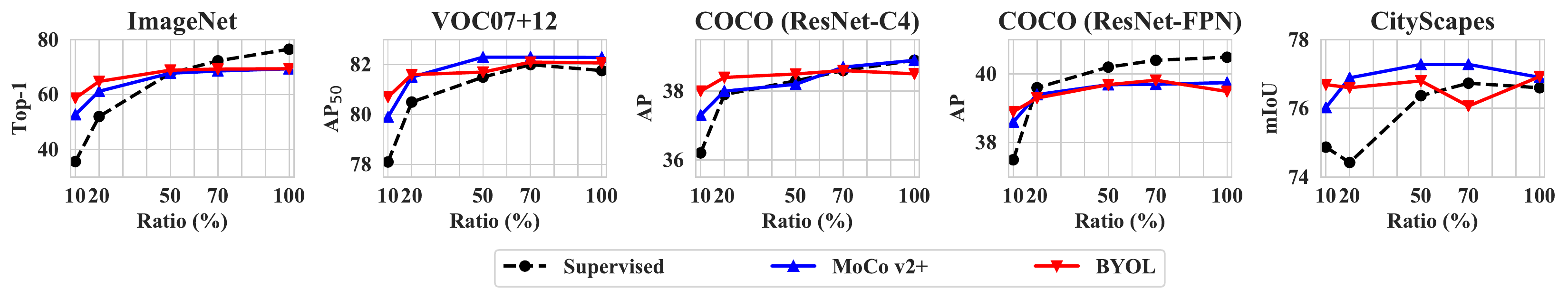}
    \caption{The results of linear evaluation and downstream tasks under different subsampling ratio}
    \label{fig:subsampling_IN}
\end{figure*}

As shown in \cref{fig:subsampling_IN}, the linear classification accuracy is monotonically related to the subsampling ratio in both supervised and self-supervised pre-training. The transfer performances reach a plateau with a relatively large subsampling ratio ($\geq$50\%). For a low-data regime ($\leq$20\%), self-supervised pre-training has a distinct advantage over supervised pre-training in transferring to downstream tasks.\\

\begin{table*}[t]
    \centering
    \caption{The performances of linear evaluation and transfer learning. All models use ResNet-50 as backbone and are trained for 100 epochs. Supervised-$X$ stands for supervised learning on ImageNet-$X$, where $X$ is the number of classes. Note that $2^{\flat}$ means class taxonomy with artifact and non-artifact, and that $2^{\diamond}$ means the class taxonomy with imbalanced data distribution}
    \scalebox{0.96}{
    \begin{tabular}{l|c|c c|c c c c|c}
    \multirow{2}{*}{} & ImageNet & VOC07 & VOC07+12 & \multicolumn{4}{c|}{COCO} & CityScapes\\
    \cline{2-9}
    & Acc & $\text{AP}_{50}$ & $\text{AP}_{50}$ & $\text{AP}_{\text{box}}^{\text{C4}}$ &
    $\text{AP}_{\text{seg}}^{\text{C4}}$ & $\text{AP}_{\text{box}}^{\text{FPN}}$ &  $\text{AP}_{\text{seg}}^{\text{FPN}}$ & mIoU\\
    \toprule
    % Random Init & -- & 33.8 & 59.6 & 29.3 & 26.4 & 32.8 & 29.9 & \\
    $\text{Supervised-}2^{\flat}$ & 9.2 & 70.2 & 79.1 & 36.8 & 32.3 & 37.6 & 34.0 & 75.7 \\
    $\text{Supervised-}2^{\diamond}$ & 0.7 & 29.2 & 46.8 & 20.9 & 19.2 & 26.3 & 24.4 & 63.6 \\
    Supervised-10 & 10.6 & 66.5 & 77.1 & 35.8 & 31.6 & 36.4 & 33.0 & 73.0 \\
    Supervised-79 & 45.4 & 74.4 & 80.9 & 38.3 & 33.6 & 39.8 & 36.0 & 76.1 \\
    Supervised-127 & 53.3 & 74.9 & 81.0 & 38.5 & 33.4 & 39.7 & 36.0 & 75.2 \\
    Supervised-1000 & 77.1 & 76.4 & 81.8 & 38.9 & 33.9 & 40.5 & 36.4 & 76.0 \\
    \midrule
    MoCo v2+ & 69.1 & 77.0 & 82.3 & 38.9 & 34.1 & 39.7 & 35.8 & 76.9 \\
	S-MoCo v2+ & 69.3 & 76.0 & 82.3 & 38.4 & 33.6 & 39.7 & 36.1 & 77.1 \\
    BYOL & 69.4 & 76.3 & 82.1 & 38.5 & 33.8 & 39.5 & 35.7 & 77.6 \\
    \end{tabular}}
    \label{tab:merging_classes}
\end{table*}

\noindent\textbf{Semantic information of labels.} Supervised pre-training with different class taxonomies in ImageNet has been discussed in \cite{imagenet_goodfor_transfer}. In this part, we study the comparison between supervised and self-supervised pre-training given different class taxonomies. The total number of training samples remains the same regardless of the change of taxonomy. We use the WordNet tree \cite{wordnet} and follow the practice of bottom-up clustering in \cite{imagenet_goodfor_transfer}, where leaf nodes belonging to the same ancestor are iteratively clustered together. According to this rule, we present four taxonomies that contain 2, 10, 79, and 127 classes respectively. We notice that the sample proportion of 2-class taxonomy is extremely imbalanced (about 1:327). To exclude the effect caused by imbalanced data distribution, we introduce another merging rule, which divides all classes into artifact and non-artifact. The data distribution is well-balanced (an approximate 1:1 ratio). \cref{tab:merging_classes} shows that when the semantic information of labels is inadequate (the number of classes is less than 79) or the data is highly imbalanced (the taxonomy with 2 imbalanced classes), self-supervised learning methods seem to be a better choice for pre-training.

\subsection{More Network Architectures}

In this subsection, we attempt to make sure that our conclusions still apply to other architectures. We adopt ResNet-18 and ResNet-101 as the backbone. We pre-train \mbox{MoCo v2}, \mbox{MoCo v2+}, and BYOL-SGD for 100 epochs. \cref{tab:other_arch} shows the linear accuracy of \mbox{MoCo v2} receives a large promotion with sophisticated model configurations ($51.9\%$ vs. $57.3\%$ for ResNet-18 and $65.0\%$ vs. $70.8\%$ for ResNet-101), which is comparable to BYOL-SGD ($57.8\%$ for ResNet-18 and $71.7\%$ on ResNet-101). Both \mbox{MoCo v2+} and \mbox{BYOL} achieve competitive results on VOC07 and VOC07+12. 

\begin{table}[h]
    \centering
    \scalebox{1.0}{
    \begin{tabular}{l|l|c|c c c|c c c}
    
     \multirow{2}{*}{Archs} & \multirow{2}{*}{Models} & ImageNet & \multicolumn{3}{c|}{VOC07} & \multicolumn{3}{c}{VOC07+12} \\
     & & Acc & $\text{AP}_{50}$ & AP & $\text{AP}_{75}$ &$\text{AP}_{50}$ &  AP & $\text{AP}_{75}$ \\
    \midrule
    
    \multirow{3}{*}{ResNet-18}
    & MoCo v2 & 51.9  & 70.3  & 40.9  & 41.0  & 78.3  & 50.2  & 54.2  \\
    & MoCo v2+ & 57.3  & 71.2  & 41.2  & 41.3  & 78.6  & 50.7  & 55.9  \\
    & BYOL-SGD & 57.8  & 71.1  & 42.0  & 43.3  & 78.7  & 51.1  & 55.4 \\
    \midrule
    
    \multirow{3}{*}{ResNet-101}
    & MoCo v2 & 65.0 & 76.7 & 50.2 & 55.1 & 82.5 & 59.3 & 65.8 \\
    & MoCo v2+ & 70.8 & 76.9 & 50.3 & 55.3 & 82.8 & 59.1 & 65.5 \\
    & BYOL-SGD & 71.7 & 76.9 & 50.2 & 55.1 & 82.6 & 59.3 & 65.6 \\
    
    \end{tabular}}
    \caption{Linear evaluation on ImageNet and detection results on VOC07 and VOC07+12 with ResNet-18 and ResNet-101.}
    \label{tab:other_arch}
\end{table}

\subsection{Other Anchors for NormRescale}
Here, we explore other anchor choices for NormRescale. We use the released supervised\footnote[1]{https://download.pytorch.org/models/resnet50-0676ba61.pth} and self-supervised (\mbox{MoCo v2}\footnote[2]{https://github.com/facebookresearch/moco}) pre-trained model as the anchor to rescale the LARS-trained BYOL. 

\begin{table}[h]
    \centering
    \scalebox{1.0}{
    \begin{tabular}{l|c c c|c c c}
    
     \multirow{2}{*}{$\mathbf{w}_{\text{S}}$} & \multicolumn{3}{c|}{VOC07} & \multicolumn{3}{c}{VOC07+12} \\
     & $\text{AP}_{50}$ & AP & $\text{AP}_{75}$ &$\text{AP}_{50}$ &  AP & $\text{AP}_{75}$ \\
    \midrule
    
    w/o rescale & 71.7 & 38.8 & 37.0 & 79.1 & 48.7 & 51.7 \\
    \midrule
    BYOL-SGD & 76.6 & 48.1 & 51.6 & 82.1 & 56.7 & 62.9 \\
    Constant & 76.1 & 47.9 & 51.8 & 82.3 & 56.8 & 62.7 \\
    MoCo v2 & 76.4 & 48.1 & 52.0 & 82.2 & 56.5 & 62.8 \\
    Supervised & 76.2 & 48.1 & 52.2 & 82.4 & 56.7 & 63.1 \\

    \end{tabular}}
    \caption{The fine-tuning results on VOC07 and VOC07+12 of LARS-trained BYOL re-scaled by different SGD-trained models or constant.}
    \label{tab:norm_rescale}
\end{table}

As we can see in \cref{tab:norm_rescale}, Using the released model like supervised or \mbox{MoCo v2} also bring about good results. Besides, we find the norm values of LARS-trained weight are roughly 10 times to that of SGD-trained weight (also shown in \mbox{Fig.~3b} in our paper), so we simply rescale the weight norm by a factor of 0.1 (abbreviated as ``Constant''). \cref{tab:norm_rescale} shows that even rescale the norm with a constant, the model does not experience significant performance degradation. NormRescale is rather robust to the choice of anchor model.

\clearpage

\end{document}